\documentclass[conference]{IEEEtran}
\IEEEoverridecommandlockouts
\usepackage{cite}
\usepackage{amsmath,amssymb,amsfonts}
\usepackage{algorithmic}
\usepackage{graphicx}
\usepackage{textcomp}
\usepackage{xcolor}
\usepackage{url}
\usepackage{multirow} 
\usepackage{tabularx} 
\def\BibTeX{{\rm B\kern-.05em{\sc i\kern-.025em b}\kern-.08em
    T\kern-.1667em\lower.7ex\hbox{E}\kern-.125emX}}
\begin{document}

\title{Research on Tibetan Tourism Viewpoints information generation system based on LLM\\

}

\author{\IEEEauthorblockN{Jinhu Qi}
\IEEEauthorblockA{\textit{Department of Artificial Intelligence
} \\
\textit{Chengdu Jincheng College}\\
Chengdu, China \\
qijinhu1218@gmail.com}
~\\
\and
\IEEEauthorblockN{Shuai Yan*}
\IEEEauthorblockA{\textit{Department of Artificial Intelligence
} \\
\textit{Chengdu Jincheng College}\\
Chengdu, China \\
yanshuai1@cdjcc.edu.cn}
*Corresponding author
~\\
\and
\IEEEauthorblockN{Wentao Zhang}
\IEEEauthorblockA{\textit{Department of Software Engineering} \\
\textit{Chengdu Jincheng College}\\
Chengdu, China \\
vraniumzwt@gmail.com}
~\\
\and
\IEEEauthorblockN{Yibo Zhang}
\IEEEauthorblockA{\textit{Department of Computer Science} \\
\textit{Chengdu Jincheng College}\\
Chengdu, China \\
z1575075389@gmail.com}

\and
\IEEEauthorblockN{Zirui Liu}
\IEEEauthorblockA{\textit{Department of Software Engineering} \\
\textit{Chengdu Jincheng College}\\
Chengdu, China \\
liuzirui733@gmail.com}

\and
\IEEEauthorblockN{Ke Wang}
\IEEEauthorblockA{\textit{Department of Artificial Intelligence
}  \\
\textit{Chengdu Jincheng College}\\
Chengdu, China \\
wangke@cdjcc.edu.cn
}
}

\maketitle

\begin{abstract}
Tibet, ensconced within China’s territorial expanse, is distinguished by its labyrinthine and heterogeneous topography, a testament to its profound historical heritage, and the cradle of a unique religious ethos. The very essence of these attributes, however, has impeded the advancement of Tibet’s tourism service infrastructure, rendering existing smart tourism services inadequate for the region’s visitors. This study delves into the ramifications of informational disparities at tourist sites on Tibetan tourism and addresses the challenge of establishing the Large Language Model (LLM) evaluation criteria. It introduces an innovative approach, the DualGen Bridge AI system, employing supervised fine-tuning techniques to bolster model functionality and enhance optimization processes. Furthermore, it pioneers a multi-structured generative results assessment framework. Empirical validation confirms the efficacy of this framework. The study also explores the application of the supervised fine-tuning method within the proprietary DualGen Bridge AI, aimed at refining the generation of tourist site information. The study’s findings offer valuable insights for optimizing system performance and provide support and inspiration for the application of LLM technology in Tibet’s tourism services and beyond, potentially revolutionizing the smart tourism industry with advanced, tailored information generation capabilities.
\end{abstract}


\section{Introduction}
\subsection{The impact of poor tourist viewpoint information on Tibet’s tourism industry}
With the rapid development of tourism information, building a smart tourism service system tailored to Tibet's unique characteristics is crucial for overcoming current deficiencies in the region's tourism infrastructure\cite{Sano2021}. This initiative is essential to promote the growth of Tibet's tourism industry. The region's unique geography, environmental conditions, and cultural background result in diverse and complex tourist preferences, necessitating a substantial amount of high-quality viewpoint information to optimize tourism services effectively.

In light of this, we embarked on the Tibet Autonomous Region Science and Technology Plan project titled "Development and Application of Tibet Intelligent Tourism Service System Based on LLM."\cite{Zhang2022}Tibet's challenging terrain, transportation difficulties, and gaps in viewpoint information hinder the full development and utilization of its tourism resources. Tourists often miss viewpoints or activities that match their preferences, leading to a perception of low tourism value and subpar services, which diminishes travel satisfaction and reduces the likelihood of return visits. This negatively impacts the sustainable development of Tibet's tourism industry.

Accurate identification of user intentions is vital for enhancing and updating viewpoint information. By analyzing tourists' search behaviors and preferences, we can better understand their needs, enabling timely updates and improvements to viewpoint information. This creates a feedback loop that enhances user experience and satisfaction. Therefore, research on reducing the information gap of scenic spots is essential for both optimizing user intention recognition algorithms and meeting the development needs of Tibet's tourism industry.

\subsection{Current status of application derivation of large-scale language models}
With the rapid implementation of Large Language Models (LLMs), exemplified by ChatGPT developed by Liu et al. in 2023, various fields have experienced significant advancements. Chafetz et al. (2024) highlighted that applications derived from LLMs exhibit strong capabilities in tasks such as text generation, question answering, and coding\cite{Chafetz2024}. However, despite LLMs' ability to generate information about Tibetan scenic spots, current models face challenges due to insufficient Tibet-specific content in training databases and the absence of a mature system tailored to this domain. Consequently, the generated content is often incomplete, inconsistently formatted, and lacks robustness, failing to meet users' practical needs.

To enhance LLMs for smart tourism applications in Tibet, it is essential to develop a comprehensive Tibetan scenic spot information generation system. This system should involve fine-tuning and parameter optimization, integrated training, and inference of LLMs. Leveraging LLMs' strengths in information synthesis, the goal is to create a system that intelligently recommends and introduces tourist viewpoints based on user intent. This approach aims to enhance the user travel experience, streamline scattered tourism resources, improve resource utilization efficiency, and ultimately promote the development and transformation of Tibet's tourism industry.

\subsection{Current status of LLM evaluation standards}
In the field of natural language processing (NLP), traditional evaluation metrics such as BLEU\cite{Papineni2002}, proposed by Papineni. in 2002, and METEOR \cite{Lavie2007}, introduced by Lavie and Agarwal in 2007, are widely used. However, these conventional metrics have limitations in specific scenarios. For instance, in the context of this study, the outputs generated by LLMs are not always natural sentences. To facilitate processing and evaluate model performance, the model outputs may be formatted in a specific structure rather than in smooth, natural language. These outputs often include geographical locations, viewpoint names, historical information, and other terms, with unnecessary connective text omitted to streamline dataset processing.

This method of generation poses challenges when using traditional coherence-based evaluation metrics, which can result in issues such as semantic dispersion and illogicality in the evaluation of Tibetan scenic spot information generation. Consequently, evaluating such outputs with standard natural sentence coherence metrics may not accurately reflect their quality or utility. Therefore, it is imperative to design a new evaluation standard that considers the specific requirements and practical application scenarios of this task. This new standard should effectively capture the multifaceted capabilities of LLMs in generating structured information for Tibetan scenic spots.

\section{Related models and Fine-tuning method}

\subsection{Supervised Fine-Tuning - SFT}
Supervised Fine-Tuning (SFT) is a fine-tuning method proposed by Ouyang\cite{Ouyang2022}. in 2022 that uses supervised learning to refine pre-trained language models. This approach employs a pre-trained language model as the base and utilizes supervised learning to adjust the model's parameters. Compared to traditional Fine-Tuning, SFT reduces the need for labeled data and improves performance by performing unsupervised learning on unlabeled data, then fine-tuning with a smaller labeled dataset. This method enhances the model's proficiency in understanding Chinese tasks and demonstrates multilingual capabilities\cite{Du2024}. For instance, Vedula\cite{Vedula2024}demonstrated the effectiveness of SFT in generating contextually relevant and coherent responses in multiple languages. This indicates that supervised fine-tuning can produce outputs more aligned with specific guidelines and user needs.

\subsection{Optimization - ORPO}
Odds Ratio Preference Optimization (ORPO) aims to improve model performance by modifying the negative log-likelihood loss function. This technique assigns a weaker penalty to rejected responses while amplifying the penalty for selected responses, thus distinguishing the generation style between preferred and non-preferred outputs during the warm-up and fine-tuning stages. According to Hong(2024), models fine-tuned with ORPO\cite{Hong2024}, such as Llama-2 and Mistral, have shown superior performance by utilizing Reinforcement Learning from Human Feedback (RLHF) or Direct Preference Optimization (DPO). ORPO improves model versatility and adaptability by integrating preference alignment directly in the supervised fine-tuning phase, simplifying fine-tuning processes, and enhancing model output alignment with user preferences.

\subsection{Low-Rank Adaptation of Large Language Models - LoRA}\label{AA}
Low-Rank Adaptation (LoRA) is a technique introduced by Hu\cite{Hu2021}for fine-tuning large language models in low-resource environments. LoRA reduces the number of trainable parameters by introducing low-rank matrices into the weight matrix of the pre-trained model, which is updated during fine-tuning. This approach maintains performance levels comparable to full fine-tuning while minimizing computational resources. Zhao\cite{Zhao2024} demonstrated that LoRA-enhanced models outperform GPT-4 on various benchmarks, particularly in scenarios requiring adaptation to specific tasks. LoRA's efficiency in parameter utilization makes it an advantageous method for combining multiple fine-tuning techniques, thereby optimizing model performance in resource-constrained settings.


\section{DualGen Bridge AI tourist viewpoint information generation system process and implementation}
Based on the above status quo and problems, this project team designed a DualGen Bridge AI scenic spot information generation system(DBA system), which fine-tuned the large language model and combined the fine-tuning method with LoRA to improve the efficiency of the SFT and ORPO fine-tuning processes. For this implementation, we used the LLaMA-Factory fine-tuning tool \cite{Zheng2024}, which is being researched and developed by Zheng in 2024.
The DBA system consists of the fine-tuning application of two large language models, the LLM location information keyword extraction model, the LLM tourist viewpoint information generation model, and a Bridge model. Its operation is divided into three processes. The first process is the location information keyword extraction part. The location information keywords in the Prompt proposed by the user are obtained through the LLM location information keyword extraction model. We will introduce the SFT+LoRA method in this part. Fine-tuning; the second process is the Bridge calculation part. The Bridge model based on longitude and latitude query is combined with the location information obtained in the first process to calculate the names of the five closest viewpoints to the location, and returns them in JSON format for use by the LLM tourist viewpoint information generation model; The third process is the LLM tourist viewpoint information generation part. The LLM viewpoint information generation model calculates the viewpoint names obtained through the Bridge model to generate introduction information for the corresponding viewpoints, including history, geography, and viewpoint information that tourists may be interested in. We will introduce this part separately. Fine-tuning of SFT+LoRA and ORPO+LoRA methods.

The overall process is shown in Figure 3-1:
\begin{figure}[ht!]
\centering
\includegraphics[width=\columnwidth]{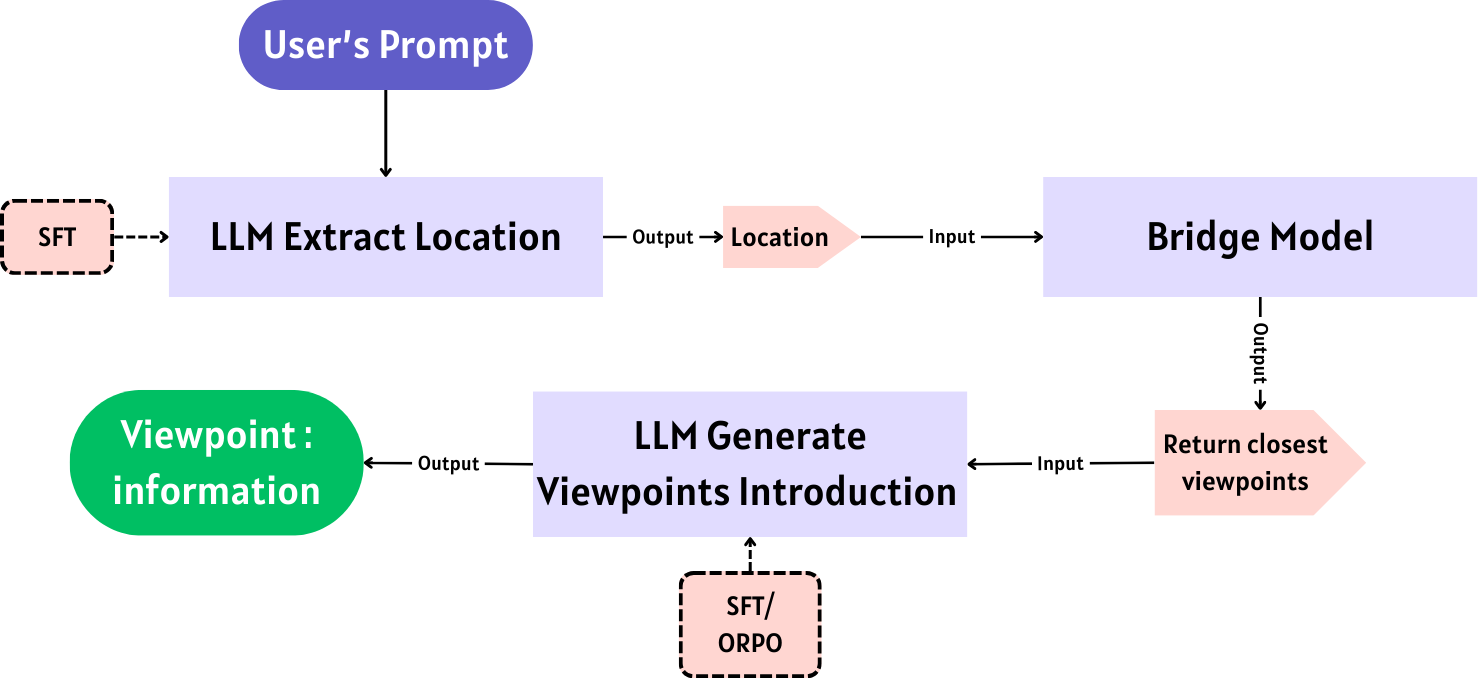} 
\caption{Figure 3-1 DualGen Bridge AI viewpoint information generation system flow chart}
\label{fig:image}
\end{figure}
\subsection{Dataset construction}
During the data acquisition phase, we adopted a multi-source data acquisition strategy to ensure the reliability of the data set. In the data acquisition stage, we obtained 421 pieces of tourist viewpoint information (including names and introductions) from Wikipedia and Ctrip, 182 pieces of hotel name information from Ctrip, 402 pieces of viewpoint latitude and longitude data in CSV format obtained through the Amap API, and The official latitude and longitude range data of the Tibet Resources Department; in the data processing stage, we first filled in and corrected the missing values of Wikipedia's erroneous tourist viewpoint information through Ctrip's tourist viewpoint information, and unified the data deduplication and formatting of all the acquired data. Standardize, and then use the longitude and latitude range data of the Tibet Resources Department to verify the longitude and latitude data of viewpoints and process missing values. Finally, the tourist viewpoint information, hotel name information, and tourist viewpoint longitude and latitude data are organized into 182 hotel information and 398 tourist viewpoint introductions. Two experiments in CSV format The hotel information data set contains the hotel name and the parsed longitude and latitude, and the scenic spot information data set contains the name, location, longitude and latitude, and a brief introduction.

\subsection{LLM location information keyword extraction model}
We are based on the idea of function calling \cite{Chen2024} to extract key information from complex text data and use the same prompt strategy to instruct the LLM location information keyword extraction model to extract location information keywords from the questions given by the user. To this end, We manually designed a set of user prompt simulation questions in various styles, selected Mistral-7B, QWEN1.5-7b, and Baichuan2-7B as prototype models for information extraction and fine-tuned the model through the SFT method. Finally, through the accuracy, The indicator scores LLM's location information keyword extraction capabilities.

\subsection{Bridge model}
The function implemented by the Bridge model is to receive the location information keywords provided by the LLM location information keyword extraction model, including place names, landmarks, or specific addresses that cannot be directly calculated. Then, by calling the Amap API, the LLM location information keyword extraction model The provided location information keywords are parsed into the longitude and latitude of the location described by the user, and then the Haversine algorithm is used to calculate the spherical distance between the location described by the user and the location of the scenic spot using the parsed longitude and latitude and the longitude and latitude in the scenic spot data set, and the nearest top- n tourist viewpoints, find the tourist viewpoint closest to the location described by the user, enter the name of the tourist viewpoint into model B, and use the introduction of the tourist viewpoint as a label. The Haversine algorithm can calculate the spherical distance between two points by calculating the difference in latitude and longitude between two points, and the sine and cosine functions of these differences. The calculation formula is as follows: 

\begin{equation} a = \sin^2\left(\frac{\Delta\phi}{2}\right) + \cos(\phi_1) \cdot \cos(\phi_2) \cdot \sin^2\left(\frac{\Delta\lambda}{2}\right) \end{equation} \begin{equation} c = 2 \cdot \arctan2\left(\sqrt{a}, \sqrt{1 - a}\right) \end{equation} \begin{equation} d = R \cdot c \end{equation}

\subsection{LLM tourist viewpoint information generation model}
The task of the LLM for generating points of interest information is to use the names of the nearest viewpoints and their related information returned by the Bridge model as inputs to generate introductory information for those viewpoints. ORPO will use the labels for preferred replies and the labels for rejected replies as fine-tuning data for further adjustments. The labels for preferred replies are defined based on the labels in the SFT fine-tuning data, while the labels for "rejected replies" are based on the responses produced by the model before fine-tuning. This method informs the LLM of preferred responses while also indicating what types of content it should not generate, thus enhancing the model's tendency toward our specified preferences and format.

\subsection{Evaluation criteria for multi-structure generation results}
We formulated a function "calculate composite score" to calculate the comprehensive score of the responses generated on the test set. This function incorporates the nonlinear weighted sum of multiple evaluation indicators. This function is designed to comprehensively evaluate and quantify the model's generation quality and performance. Below is the formula for this function to calculate a composite score to evaluate the quality of generated content for each model.\\

Comprehensive score (percentage system) = 
\begin{equation}
\begin{cases} 
a & \sum \left(c_1 \cdot \text{BLEU} + c_2 \cdot \text{Rouge-1} + \right.\\
  & \left. c_3 \cdot \text{Rouge-2} + c_4 \cdot \text{Rouge-L} + c_5 \cdot \text{accurate rate}^2\right) \\
b & \sum \left(d_1 \cdot \text{fluency } + d_2 \cdot \log(\text{accurate rate} + 1) + \right.\\
  & \left. d_3 \cdot \exp(\text{relevance })\right)
\end{cases}
\end{equation}
The method of applying this scoring formula is that when the LLM generation is structural content, the parameters of a are set to 1 and the parameters of b are set to 0 to use BLEU, rouge set, and accuracy to calculate the comprehensive score, because these indicators are calculated by The quality of the output is evaluated by the overlap (such as n-gram matching) between the text generated by the model and a set of reference texts. When LLM generates unstructured content, we need to set the parameters of a to 0 and the parameters of b to 1. The generated content needs to be compared with the reference text for semantic and sentence meaning recognition and evaluation. Assessment of semantics and sentence meaning requires the use of scores in the fluency, accuracy, and relevance domains. The fluency domain and relevance domain are based on Wang's fine-tuned Llama3-8B-Chinese-Chat \cite{Wang2024} to evaluate the correlation between each sentence and the reference sentence and the score of the fluency of the generated sentence. The score in the accurate field is the score evaluated based on BERTscore as a reference. \cite{Zhang2019} BERTScore is an evaluation tool based on the BERT model developed by Zhang. It effectively evaluates semantics by calculating the cosine similarity between the word vectors in the generated text and the reference text. Accuracy and consistency.

c1, c2, c3, c4, c5, d1, d2, d3 are the weight coefficients of different evaluation indicators, which can be adjusted according to the needs of actual evaluation. $Accuracy^2$, log(accuracy score+1), and exp(correlation) provide non-linear evaluation methods, increasing the complexity and sensitivity of scoring. a and b act as switch parameters to ensure that the correct scoring criteria are selected depending on the output type (structured or unstructured).

\section{Experiment}
\subsection{Validity experiment of scoring criteria}
In order to ensure the effectiveness of the scoring system, we selected Qwen1.5-7B-Chat as a pre-training model based on the 3.5 multi-structure generation result evaluation standard to conduct the scoring standard validity experiment. First, we constructed a task-specific dataset related to idiom extraction and interpretation to correspond to the key extraction and content generation tasks in this article. \cite{Bai2024} The data comes from COIG-CQIA released by Bai in 2024 and has been scaled. The dataset contains 111 idiom-related instances, with the unstructured portion of the test set being manually scored to assess the fluency, accuracy, and relevance of the generated text. Next, we normalized the original scores of the evaluation system to ensure that all scores were within the range of 0 to 100, eliminating the impact of dimensional differences. The score distribution results are shown in the figure below:
\begin{figure}[ht!]
\centering
\includegraphics[width=\columnwidth]{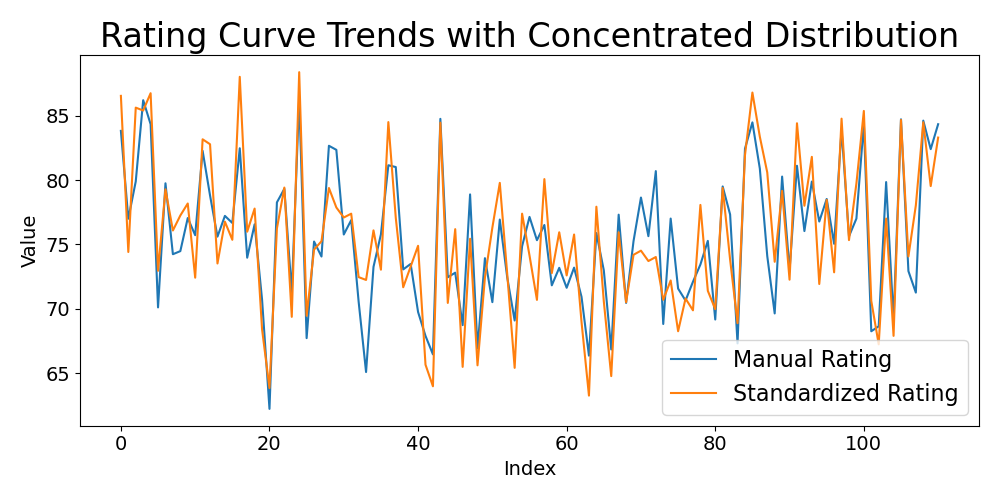} 
\caption{Data distribution chart of system scoring and manual scoring}
\label{fig:image}
\end{figure}

The results of the data analysis show that the scoring results of the scoring criteria we designed and the results of manual scoring have similar distribution patterns in the data distribution, which proves the effectiveness of the scoring system. When evaluating multi-structured data generation tasks, this system can provide feedback that is close to manual evaluation. Therefore, this scoring system can be used as a powerful tool to assist model selection and optimization decisions, ensuring the quality and applicability of generated content.

\subsection{Experiment on keyword extraction effect based on LLM location information}
The instruction of the large language model is to "parse the input text, identify and return the names of entities in it that are marked as places, ignore any other kind of information, and only extract and return the place names. Regarding prompts, we first tested based on the user's accommodation for hotel location data, the standard format of the prompt is "Please recommend me viewpoints near {hotel name}". To avoid the possibility of overfitting due to the uniform prompt format while simulating the different input habits of different users, we used a variety of prompt formats,  and the output format was unified as "{hotel name}".

First, in terms of the Supervised fine-tuning (SFT) fine-tuning method, we used 183 hotel data split into 80\% fine-tuning data and 20\% test data. Fine-tuning the data as a basis created 146 entries. Instruction is "Parse the input text, identify and return the names of entities in it that are marked as places. Ignore any other type of information, only extract and return place names", the input is "Please recommend to me viewpoints near {hotel name}", The output is "{hotel name}". In the fine-tuning parameters of SFT, we set the fine-tuning type to LoRa to improve fine-tuning efficiency and set the fine-tuning precision to FP16 to reduce memory requirements and increase computing speed. The criterion for LLM extraction of keywords is to check the matching accuracy between the label and prediction. The comprehensive score of Mistral-7B-Instruct-v0.2 is the highest among the three models, which shows that this model is very good at generating positional keyword extraction content. The best effect. Below are the domain scores and overall scores of each model.

\begin{table}[h!]
  \centering
  \caption{Model Performance Comparison}
  \label{tab:table1}
  \begin{tabular}{|c|c|c|}
    \hline
    \textbf{Model} & \textbf{Method} & \textbf{Accuracy Rate} \\
    \hline
    Mistral-7B-Instruct-v0.2 & Zero-Shot & 0.060 \\
    \cline{2-3}
     & SFT & 1.000 \\
    \hline
    Qwen1.5-7B-Chat & Zero-Shot & 0.470 \\
    \cline{2-3}
     & SFT & 0.980 \\
    \hline
    Baichuan2-7b & Zero-Shot & 0.067 \\
    \cline{2-3}
     & SFT & 0.163 \\
    \hline
  \end{tabular}
\end{table}

\begin{figure}[ht!]
\centering
\includegraphics[width=1\columnwidth]{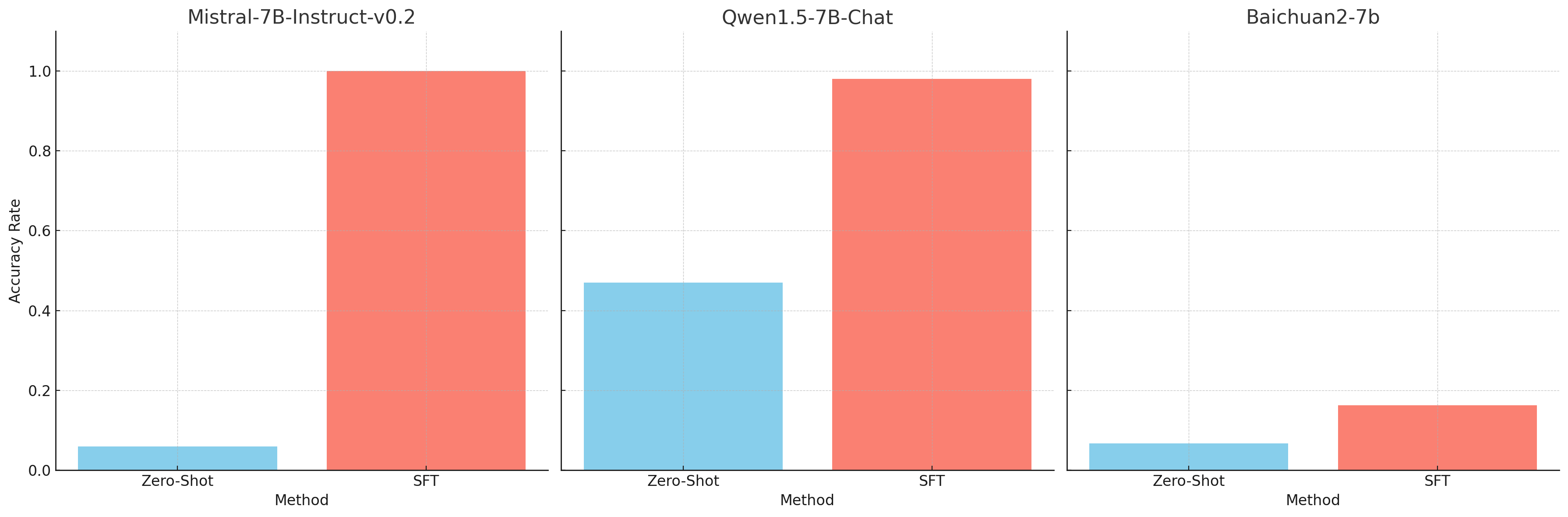} 
\caption{Zero-Shot vs SFT for Models}
\label{fig:image}
\end{figure} 

The tables and charts show the performance of each model without fine-tuning (Zero-Shot) and after fine-tuning (Supervised Fine-Tuning, SFT). The Mistral-7B-Instruct-v0.2 model has a weak ability to extract location keywords without training, with a score of only 0.10294. After using the SFT fine-tuning method, the model's performance reaches perfection, with a score of 1.00. Before Qwen1.5-7B-Chat was trained, the performance of the model was medium, with a score of 0.5007. After SFT training, the model's performance improved significantly, with a score of 0.9751. The Baichuan n2-7 model performed poorly without training, with a score of 0.0662. Even after SFT training, the model's performance on the task of extracting positional keywords is still low, with a score of 0.1353. It can be seen that after specialized fine-tuning, most models perform better on the task of extracting positional keywords in user prompts, especially Mistral-7B-Instruct-v0.2 and Qwen1.5-7B-Chat There is a higher effect improvement. Experimental results show that Mistral-7B-Instruct-v0.2 has the most outstanding performance, so in the DBA system, we will use Mistral-7B-Instruct-v0.2 fine-tuned by SFT to extract positional keywords in user prompts.

\subsection{LLM model optimization and universality comparison experiment}
Regarding the selection of the LLM scenic spot information generation model in the DBA system, we adjust its calculation accuracy parameters and use the same method to fine-tune each LLM, and then evaluate the accuracy and tuning method of the corresponding model according to the evaluation rules in 3.5. Based on the scores below, the configuration group with the best generation effect of the LLM scenic spot information generation model is finally obtained.

When using the SFT method to fine-tune LLMs to learn all relevant information about tourist viewpoints, so as to more accurately generate the viewpoints and their descriptions needed by tourists, we used all 398 viewpoint descriptions as the basis for fine-tuning. The format of `instruction` and `input` in ORPO fine-tuning is consistent with SFT, but the `output` needs a different approach to ensure that LLMs return the expected `label`, which is the viewpoint descriptions mentioned in the training data. `Output` data should be in list form, where the first element is the `chosen answer`, matching the descriptions of the 398 viewpoints, and the second is the `rejected answer`, which is the viewpoint description directly returned by the model before fine-tuning. By changing 20\% of the data to all original data, we can obtain the descriptions returned by the model before fine-tuning for each viewpoint.

\begin{table*}[t]
  \centering
  \caption{Detailed Model Performance Analysis}
  \label{tab:table1}
  \begin{tabular}{cccccccc}
    \hline
    \textbf{Model} & \textbf{Method} & \textbf{Compute type} & \textbf{fluency} & \textbf{accurate rate} & \textbf{relevance} & \textbf{Overall Score} & \textbf{Overall Score \%} \\
    \hline
    & sft & fp16 & 0.8090 & 0.6751 & 0.6493 & 1.1631 & 72.91 \\
             &     & fp32 & 0.8094 & 0.6795 & 0.6462 & 1.1617 & 72.82 \\
       llam3-8b       &     & bf16 & 0.8150 & 0.6787 & 0.6743 & 1.1849 & 74.28 \\
             & orpo & fp16 & 0.8150 & 0.6667 & 0.6551 & 1.1678 & 73.21 \\
             &     & fp32 & 0.8061 & 0.6620 & 0.6374 & 1.1509 & 72.15 \\
             &     & bf16 & 0.8056 & 0.6610 & 0.6221 & 1.1390 & 71.40 \\
\hline
    & sft & fp16 & 0.7943	& 0.6632 & 0.64675 & 1.1546 & 72.38 \\
             &     & fp32 & 0.7837&	0.6681&	0.6646&	1.1661&	73.10\\
       qwen1.5-7b      &     & bf16 & 0.8018	&0.6680	&0.6924	&1.1934	&74.81 \\
             & orpo & fp16 &0.7737	&0.6495&	0.6285&	1.1321	&70.97 \\
             &     & fp32 & 0.7650	&0.6579&	0.6329&	1.1344&	71.11 \\
             &     & bf16 & 0.7933&0.6529	&0.6331&	1.1421&	71.60 \\
 \hline 
     & sft & fp16 & 0.5734	&0.6074&	0.5485&	1.0067	&63.11\\
             &     & fp32 & 0.5655	&0.6037&	0.5392&	0.9971&	62.51 \\
      baichuan 2-7b       &     & bf16 & 0.5346&	0.6019&	0.5285&	0.9803&	61.45\\
             & orpo & fp16 & 0.5213&	0.5990&	0.5081&	0.9621&	60.31 \\
             &     & fp32 & 0.5361	&0.6011&	0.5484&	0.9942&	62.32\\
             &     & bf16 & 0.5246&	0.6018&	0.5241&	0.9743&	61.08 \\
  \hline
     & sft & fp16 & 0.8436	&0.6857&	0.7796&	1.2819&	80.36 \\
             &     & fp32 & 0.8325&	0.6879&	0.7451&	1.2495&	78.33\\
      chatglm3-6b       &     & bf16 & 0.8450&	0.6825&	0.7885&	1.2896&	80.84 \\
             & orpo & fp16 &0.8523&	0.6790&	0.7754&	1.2797&	80.22\\
             &     & fp32 & 0.8158&	0.6794&	0.7259&	1.2269&	76.91\\
             &     & bf16 & 0.8358	&0.6832&	0.7821&	1.2814&	80.33 \\
    \hline
  \end{tabular}
\end{table*}

\begin{figure*}
    \centering
    \includegraphics[width=\textwidth,height=8cm]{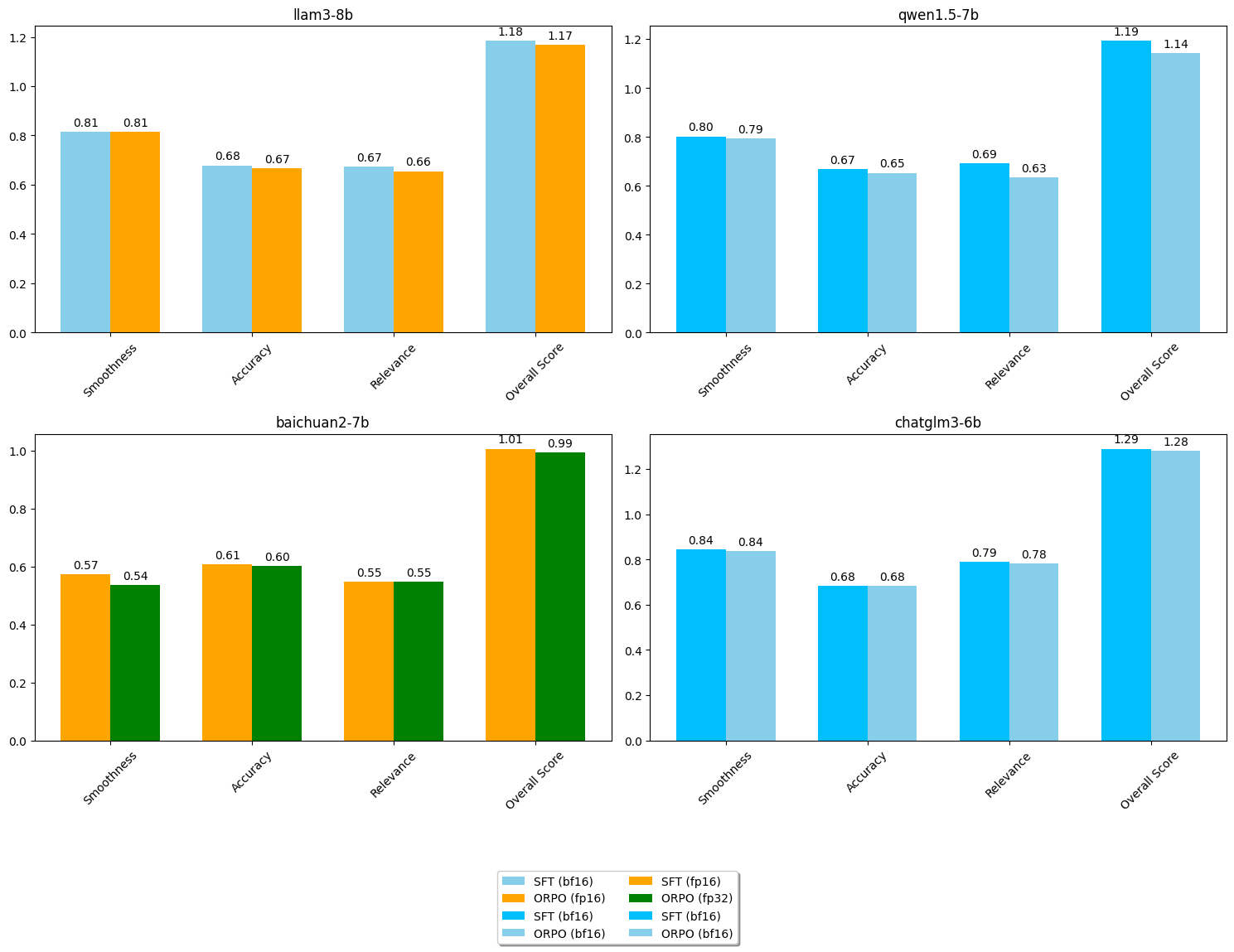}
    \caption{SFT vs ORPO in Different Computing Type for Models}
\end{figure*}

It can be seen from the experimental results that the Llama3-8b-Chat model has the best quality of generated content under the BF16 precision parameter of the SFT fine-tuning method, with a comprehensive score of 74.28. The qwen1.5-7b model also performs under the BF16 precision parameter under the SFT fine-tuning method. The quality of generated content is the best, with a comprehensive score of 74.81. The Baichuan2-7b model has the best quality of generated content under the SFT fine-tuning method under the FP16 accuracy parameter, with a comprehensive score of 63.11. Under the SFT fine-tuning method, the ChatGLM3-6b model with an accuracy of BF16 produced the best quality content with an overall score of 80.84. From the comparison of the comprehensive scores of each model, the ChatGLM3-6b model has the most outstanding effect under the SFT fine-tuning method BF16 precision parameters. The smoothness, accuracy, and correlation of the model The sex score is also very high compared to other models. Its fluency is second in this field, its accuracy is third in this field, and its correlation score is first in this field. This can explain the ChatGLM3-6b model The SFT fine-tuning method BF16 accuracy parameter is relatively more suitable for the configuration needs of the LLM scenic spot information generation model.

\section{Conclusion}
Through this research, we have developed the DualGen Bridge AI viewpoints Information Generation System, which is oriented towards the research topic "Development and Application of LLM-based Intelligent Tourism Service System for Tibet." In this project, we have proposed and implemented for the first time a dual-model architecture called the DualGen Bridge AI viewpoints Information Generation System, combining positional keyword extraction and viewpoints information generation. This architecture significantly improves the accuracy and diversity of the generated information, providing new principles and experimental evidence for future research in the same field.

For the system evaluation task, we proposed a multi-structure evaluation standard for the generated results and validated its effectiveness in experiments, filling the gap in the field's lack of a unified standard. This provides theoretical and experimental evidence for future research on the evaluation of large models and information generation, specifically in terms of assessing the diversity and accuracy of the generated results.

Based on the above work, we validated the effectiveness of using LoRA technology for efficient parameter fine-tuning under specific parameters by combining LoRA technology with SFT/ORPO. This approach retains the original performance of large models while improving their performance on specific tasks, optimizing computational costs and adaptability. This provides new directions and experimental evidence for the accuracy parameter optimization and application scenario selection of large-scale language models in the future, emphasizing the importance of fine-tuning methods and accuracy parameter selection in model performance optimization.

Ultimately, we verified the effectiveness of extracting positional keywords from user prompts and generating viewpoints content introduction information and concluded the optimal model configuration for the DualGen Bridge AI viewpoints Information Generation System: the SFT fine-tuned LLM positional information keyword extraction model is the 7 billion parameter v2 model released by Mistral, and the viewpoints information generation model is the 6 billion parameter ChatGLM3 model released by Tsinghua University in BF16 computational precision mode.

In the future, we will strive to collect more dimensional information and attempt to incorporate multi-modal tasks into this system architecture, further enhancing the generalization capability of the DBA system to meet more diverse user needs.





\vspace{12pt}


\begin{thebibliography}{00}

\bibitem{Sano2021}
K. Sano and H. Sano, "Value co-creation and open innovation in urban smart tourism ecosystem development: A lens from service dominant logic," 2021.

\bibitem{Zhang2022}
R. Zhang, et al., "Construction and countermeasure analysis of Tibet's all-region smart tourism ecosystem platform," \textit{Journal of North China University of Science and Technology (Social Sciences Edition)}, vol. 22, no. 4, pp. 27-32+41, 2022.

\bibitem{Chafetz2024}
H. Chafetz, S. Saxena, and S. G. Verhulst, "A Fourth Wave of Open Data? Exploring the Spectrum of Scenarios for Open Data and Generative AI," 2024, \url{https://doi.org/10.48550/arXiv.2405.04333}.

\bibitem{Papineni2002}
K. Papineni, S. Roukos, T. Ward, and W. J. Zhu, "BLEU: A method for automatic evaluation of machine translation," in \textit{Proceedings of the 40th Annual Meeting of the Association for Computational Linguistics}, 2002, pp. 311-318.

\bibitem{Lavie2007}
A. Lavie and A. Agarwal, "METEOR: An automatic metric for MT evaluation with improved correlation with human judgments," 2007, \url{https://doi.org/10.3115/1626355.1626389}.

\bibitem{Ouyang2022}
L. Ouyang, et al., "Training language models to follow instructions with human feedback," arXiv preprint, 2022, \url{https://arxiv.org/abs/2203.02155}.

\bibitem{Du2024}
X. Du, et al., "Chinese Tiny LLM: Pretraining a Chinese-Centric Large Language Model," arXiv preprint, 2024, \url{https://arxiv.org/abs/2404.04167}.

\bibitem{Vedula2024}
N. Vedula, O. Rokhlenko, and S. Malmasi, "Question Suggestion for Conversational Shopping Assistants Using Product Metadata," 2024, \url{https://doi.org/10.1145/3626772.3661371}.

\bibitem{Hong2024}
J. Hong, N. Lee, and J. Thorne, "ORPO: Monolithic Preference Optimization without Reference Model," 2024, \url{https://doi.org/10.48550/arXiv.2403.0769}.

\bibitem{Hu2021}
E. J. Hu, Y. Shen, P. Wallis, Y. Li, S. Wang, and W. Chen, "LoRA: Low-Rank Adaptation of Large Language Models," 2021, \url{arXiv:2106.09685}.

\bibitem{Zhao2024}
J. Zhao, T. Wang, W. Abid, G. Angus, A. Garg, J. Kinnison, A. Sherstinsky, P. Molino, T. Addair, D. Rishi, and others, "LoRA Land: 310 Fine-tuned LLMs that Rival GPT-4," 2024, \url{arXiv:2405.00732}.

\bibitem{Zheng2024}
Y. Zheng, R. Zhang, J. Zhang, Y. Ye, Z. Luo, and Y. Ma, "LlamaFactory: Unified Efficient Fine-Tuning of 100+ Language Models," arXiv preprint, 2024, \url{https://arxiv.org/abs/2403.13372}.

\bibitem{Chen2024}
W. Chen, Z. Li, and M. Ma, "Octopus: On-device language model for function calling of software APIs," arXiv preprint, 2024, \url{https://arxiv.org/abs/2404.01549}.

\bibitem{Wang2024}
S. Wang and Y. Zheng, "Llama3-8B-Chinese-Chat," \url{https://huggingface.co/shenzhi-wang/Llama3-8B-Chinese-Chat}.

\bibitem{Zhang2019}
T. Zhang, V. Kishore, F. Wu, K. Q. Weinberger, and Y. Artzi, "BERTScore: Evaluating Text Generation with BERT," 2019, \url{https://arxiv.org/abs/1904.09675}.

\bibitem{Bai2024}
Y. Bai, X. Du, Y. Liang, Y. Jin, Z. Liu, J. Zhou, T. Zheng, X. Zhang, N. Ma, Z. Wang, et al., "COIG-CQIA: Quality is All You Need for Chinese Instruction Fine-Tuning," arXiv preprint, 2024, \url{https://arxiv.org/abs/2403.18058}.

\end{thebibliography}
\end{document}